\documentclass{article}
\usepackage{spconf,amsmath,graphicx}
\usepackage{color}
\usepackage{url}

\DeclareMathOperator*{\argmax}{argmax}

\title{LEVERAGING NATIVE LANGUAGE SPEECH FOR ACCENT IDENTIFICATION USING DEEP SIAMESE NETWORKS}
\name{Aditya Siddhant$^{\dagger}$, Preethi Jyothi$^{\S}$, Sriram Ganapathy$^{\ddagger}$\sthanks{This work was carried out with the help of a research grant awarded by Microsoft Research India (MSRI) for the Summer Workshop on Artificial Social Intelligence.}}
\address{$^{\dagger}$Carnegie Mellon University, Pittsburgh, PA, USA\\
$^{\S}$Indian Institute of Technology Bombay, Mumbai, India \\
$^{\ddagger}$Indian Institute of Science, Bengaluru, India}
\begin{document}
%
\maketitle
\begin{abstract}
The problem of automatic accent identification is important for several applications like speaker profiling and recognition as well as for improving speech recognition systems. The accented nature of speech can be primarily attributed to the influence of the speaker's native language on the given speech recording. In this paper, we propose a novel accent identification system whose training exploits speech in native languages along with the accented speech. Specifically, we develop a deep Siamese network based model which learns the association between accented speech recordings and the native language speech recordings. The Siamese networks are trained with i-vector features extracted from the speech recordings using either an unsupervised Gaussian mixture model (GMM) or a supervised deep neural network (DNN) model. We perform several accent identification experiments using the CSLU Foreign Accented English (FAE) corpus. In these experiments, our proposed approach using deep Siamese networks yield significant relative performance improvements of 15.4\% on a 10-class accent identification task, over a baseline DNN-based classification system that uses GMM i-vectors. Furthermore, we present a detailed error analysis of the proposed accent identification system. 
\end{abstract}

\section{Introduction}
\label{sec:intro}

Over the recent years, many of voice-driven technologies have achieved significant robustness needed for mass deployment.
 This is largely due to significant advances in automatic speech
recognition (ASR) technologies and deep learning algorithms. However, the variability in speech accents pose
a significant challenge to state-of-the-art speech systems.  In
particular, large sections of the English-speaking population in the world
face difficulties interacting with voice-driven agents in English due to
the mis-match in speech accents seen in the training data. The accented nature of speech can be primarily
attributed to the influence of the speaker's native language. In this work
we focus on the problem of {\em accent identification}, where the user's
native language is automatically determined from their non-native speech. This
can be viewed as a first step towards building accent-aware voice-driven
systems. 

Accent identification from non-native speech bears resemblance to the task
of language identification~\cite{Zissman2001}. However, accent
identification is a harder task as many cues about the speaker's native
language are lost or suppressed in the non-native speech. Nevertheless,
one may expect that the speaker's native language is reflected in the
acoustics of the individual phones used in non-native language speech, along with
pronunciations of words and grammar. In this work, we focus on the
acoustic characteristics of an accent induced by a speaker's native
language. 

\vspace{1em}
\noindent \textbf{Our main contributions:}
\begin{itemize}
\item We develop a novel deep Siamese network based model which
learns the association between accented speech and native
language speech.
\item We explore i-vector features extracted using both an unsupervised
Gaussian mixture model (GMM) and a supervised deep neural network (DNN)
model.
\item We present a detailed error analysis of the proposed system which
reveals that the confusions among accent predictions are contained within the language family of the corresponding native language. 
\end{itemize}

Section~\ref{sec:factor_analysis} outlines the i-vector feature extraction
process. Section~\ref{sec:approach} describes our Siamese network-based
model for accent identification. Our experimental results are detailed in
Section~\ref{sec:expts} and Section~\ref{sec:discussion} provides an error
analysis of our proposed approach.

\section{Related Work}
\label{sec:relatedwork}

The prior work on foreign accent identification has drawn inspiration from
techniques used in language identification~\cite{Zissman1996}. The 
phonotactic model based approaches~\cite{Biadsy2011} and acoustic model based
approaches~\cite{Teixeira1996} have been explored for accent identification
in the past. More recently, i-vector based representations,  which is part of the
state-of-the-art speaker recognition~\cite{Dehak2011} and
language recognition~\cite{Dehak2011b} systems, have been applied to the task of
accent recognition. The i-vector systems that used GMM-based background models were
found to outperform other competitive baseline
systems~\cite{Bahari2013,Lazaridis2014,Najafian2016}.

In the recent years, the language recognition systems and speaker recognition systems have shown promising results with the use of deep neural network (DNN) model based i-vector extraction \cite{lei,ibm}. 
Also, none of the previous approaches have exploited speech
in native languages while training accent identification systems to the best of our knowledge. This work attempts to
develop accent recognition systems using both these components.

\section{Factor Analysis Framework for i-vector Extraction}
\label{sec:factor_analysis}
The techniques outlined here are derived from previous work on joint
factor analysis (JFA) and i-vectors \cite{kenny,kenny_jfa,najim_spk}. We
follow the notations used in \cite{kenny}. The training data from all the
speakers is used to train a GMM with model parameters $\lambda = \{ \pi_c ,
\boldsymbol{\mu_c}, \boldsymbol{\Sigma_c} \}$ where $\pi_c$,
$\boldsymbol{\mu}_c$ and $\boldsymbol{\Sigma}_c$ denote the mixture
component weights, mean vectors and covariance matrices respectively for
$c=1,..,C$ mixture components. Here, $\boldsymbol{\mu}_c$  is a vector of
dimension $F$ and $\boldsymbol{\Sigma}_c$ is of assumed to be a diagonal
matrix of dimension $F\times F$.
\subsection{GMM-based i-vector Representations}\label{sec:ivec}
Let $\boldsymbol{\mathcal{M}_0}$ denote the universal background model (UBM) supervector which is the
concatenation of $\boldsymbol{\mu}_c$ for $c=1,..,C$ and is of dimensionality 
$D \times 1$ (where $D = C \cdot F$). Let $\mathcal{\boldsymbol{\Sigma}}$ denote the block diagonal
matrix of size $D \times D$ whose diagonal blocks are
$\boldsymbol{\Sigma}_c$. Let
$\mathcal{X}(s)=\{\it{\boldsymbol{x}}_i^s,i=1,...,H(s)\}$ denote the
low-level feature sequence for input recording $s$ where $i$ denotes the
frame index. Here $H(s)$ denotes the number of frames in the recording. Each
$\it{\boldsymbol{x}}_i^s$ is of dimension $F \times 1$.

Let $\boldsymbol{\mathcal{M}(s)}$ denote the recording supervector which is
the concatenation of speaker adapted GMM means $\boldsymbol{\mu}_c(s)$ for
$c=1,..,C$ for the speaker $s$. Then, the i-vector model is,
\begin{equation} \label{eq:fa_model}
\boldsymbol{\mathcal{M}}(s) = \boldsymbol{\mathcal{M}_0} + \boldsymbol{V}\boldsymbol{y}(s)
\end{equation}
where $\boldsymbol{V}$ denotes the total variability matrix of dimension $D
\times M$ and $\boldsymbol{y}(s)$ denotes the i-vector of dimension $M$. The
i-vector is assumed to be distributed as
$\mathcal{N}(\boldsymbol{0},\boldsymbol{I})$.

In order to estimate the i-vectors, the iterative EM algorithm is used. We
begin with a random initialization of the total variability matrix
$\boldsymbol{V}$. Let $p_{\lambda}(c | {\boldsymbol{x}}_i^s)$ denote the
alignment probability of assigning the feature vector ${\boldsymbol{x}}_i^s$
to mixture component $c$. The sufficient statistics are then computed as,
\begin{equation} \label{eq:stats}
\begin{split}
N_c(s) & = \sum _{i=1}^{H(s)} p_{\lambda}(c | {\boldsymbol{x}}_i^s) \\
\boldsymbol{S}_{X,c}(s) & = \sum _{i=1}^{H(s)} p_{\lambda}(c | {\boldsymbol{x}}_i^s) ({\boldsymbol{x}}_i^s - \boldsymbol {\mu} _c)
\end{split}
\end{equation}
Let $\boldsymbol{N(s)}$ denote the $D\times D$ block diagonal matrix with
diagonal blocks $N_1(s)\boldsymbol{I}$,
$N_2(s)\boldsymbol{I}$,..,$N_C(s)\boldsymbol{I}$ where $\boldsymbol{I}$ is
the $F\times F$ identity matrix. Let $\boldsymbol{S}_X(s)$ denote the
$D\times 1$ vector obtained by splicing
$\boldsymbol{S}_{X,1}(s)$,..,$\boldsymbol{S}_{X,C}(s)$.

It can be easily shown~\cite{kenny} that the posterior distribution of the
i-vector $p_{\lambda}(\boldsymbol{y}(s) | \mathcal{X}(s))$ is Gaussian with
covariance ${\boldsymbol{\mathnormal{l}}}^{-1}(s)$ and mean
${\boldsymbol{\mathnormal{l}}}^{-1}(s)\boldsymbol{V}^*{\mathcal{\boldsymbol{\Sigma}}}^{-1}\boldsymbol{S}_X(s)$,
where
\begin{equation} \label{eq:l(s)}
\boldsymbol{\mathnormal{l}}(s) = \boldsymbol{I} + \boldsymbol{V}^*{\mathcal{\boldsymbol{\Sigma}}}^{-1}\boldsymbol{N}(s)\boldsymbol{V}
\end{equation}
The optimal estimate for the i-vector $\boldsymbol{y}(s)$ obtained as
$\argmax_y \big [ p_{\lambda}(\boldsymbol{y}(s) | \mathcal{X}(s)) \big ]$ is
given by the mean of the posterior distribution.

For re-estimating the $\boldsymbol{V}$ matrix, the maximization of the
expected value of the log-likelihood function (EM algorithm), gives the
following relation \cite{kenny},
\begin{equation} \label{eq:mstep}
\sum _{s=1}^{S} \boldsymbol{N}(s)~ \boldsymbol{V}~ \mathrm{E} \big[\boldsymbol{y}(s) \boldsymbol{y}^*(s) \big ] = \sum _{s=1}^{S} \boldsymbol{S}_X(s)  \mathrm{E} \big [\boldsymbol{y}^*(s) \big ]
\end{equation}
where $\mathrm{E}[.]$ denotes the posterior expectation operator. The
solution for Eq.~(\ref{eq:mstep}) can be computed for each row of
$\boldsymbol{V}$. Thus, the i-vector estimation is performed by iterating
between the estimation of posterior distribution  and the update of the
total variability matrix (Eq.~(\ref{eq:mstep})).

\subsection{DNN i-vectors} \label{sec:dnn_ivec}

Instead of using a GMM-UBM based computation of i-vectors, we can also use
DNN-based context dependent state (senone) posteriors to generate the sufficient
statistics used in the i-vector computation \cite{mclaren,lei}.  The GMM
mixture components will be replaced with the senone classes present at the
output of the DNN. Specifically, $p_{\lambda}(c | {\boldsymbol{x}}_i^s)$
used in Eq.~(\ref{eq:stats}) is replaced with the DNN posterior probability
estimate of the senone $c$ given the input acoustic feature vector
$\boldsymbol{x}_i^s$ and the total number of senones is the parameter $C$. The
other parameters of the UBM model  $\lambda = \{ \pi_c , \boldsymbol{\mu_c},
\boldsymbol{\Sigma_c} \}$  are computed as

\begin{equation} \label{eq:dnn_ivec}
\begin{split}
\pi_c & = \frac {\sum _{s=1}^{S} \sum_{i=1}^{H(s)} p(c | {\boldsymbol{x}}_i^s)} {\sum_{c=1}^C\sum _{s=1}^{S} \sum_{i=1}^{H(s)}p(c | {\boldsymbol{x}}_i^s) } \\
\boldsymbol{\mu}_c & = \frac {\sum _{s=1}^{S} \sum_{i=1}^{H(s)} p(c | {\boldsymbol{x}}_i^s) \boldsymbol{x}_i^s } {\sum_{c=1}^C\sum _{s=1}^{S} \sum_{i=1}^{H(s)}p(c | {\boldsymbol{x}}_i^s) } \\
\boldsymbol{\Sigma}_c & = \frac {\sum _{s=1}^{S} \sum_{i=1}^{H(s)} p(c | {\boldsymbol{x}}_i^s) (\boldsymbol{x}_i^s - \boldsymbol{\mu}_c) (\boldsymbol{x}_i^{s}-\boldsymbol{\mu}_c)^*} {\sum_{c=1}^C\sum _{s=1}^{S} \sum_{i=1}^{H(s)}p(c | {\boldsymbol{x}}_i^s) }
\end{split}
\end{equation}

\section{Proposed Approach}
\label{sec:approach}

Siamese networks~\cite{Bromley1994,Chopra2005} are neural network models
that are designed to learn a similarity function between pairs of input representations. This architecture consists of two identical neural networks with shared weights where the input is a pair of samples. The objective is to find a function that minimizes or maximizes the similarity between the pair of inputs depending on whether they belong to the same category or not. This is achieved by optimizing a contrastive loss function containing dual terms - a component that reduces the contribution from the positive training examples (i.e. pairs of inputs belonging to the same category) and a component that increases the contribution from negative training examples (i.e. pairs of inputs from different categories).
\begin{figure}[t!]
\centering
\includegraphics[width=\linewidth]{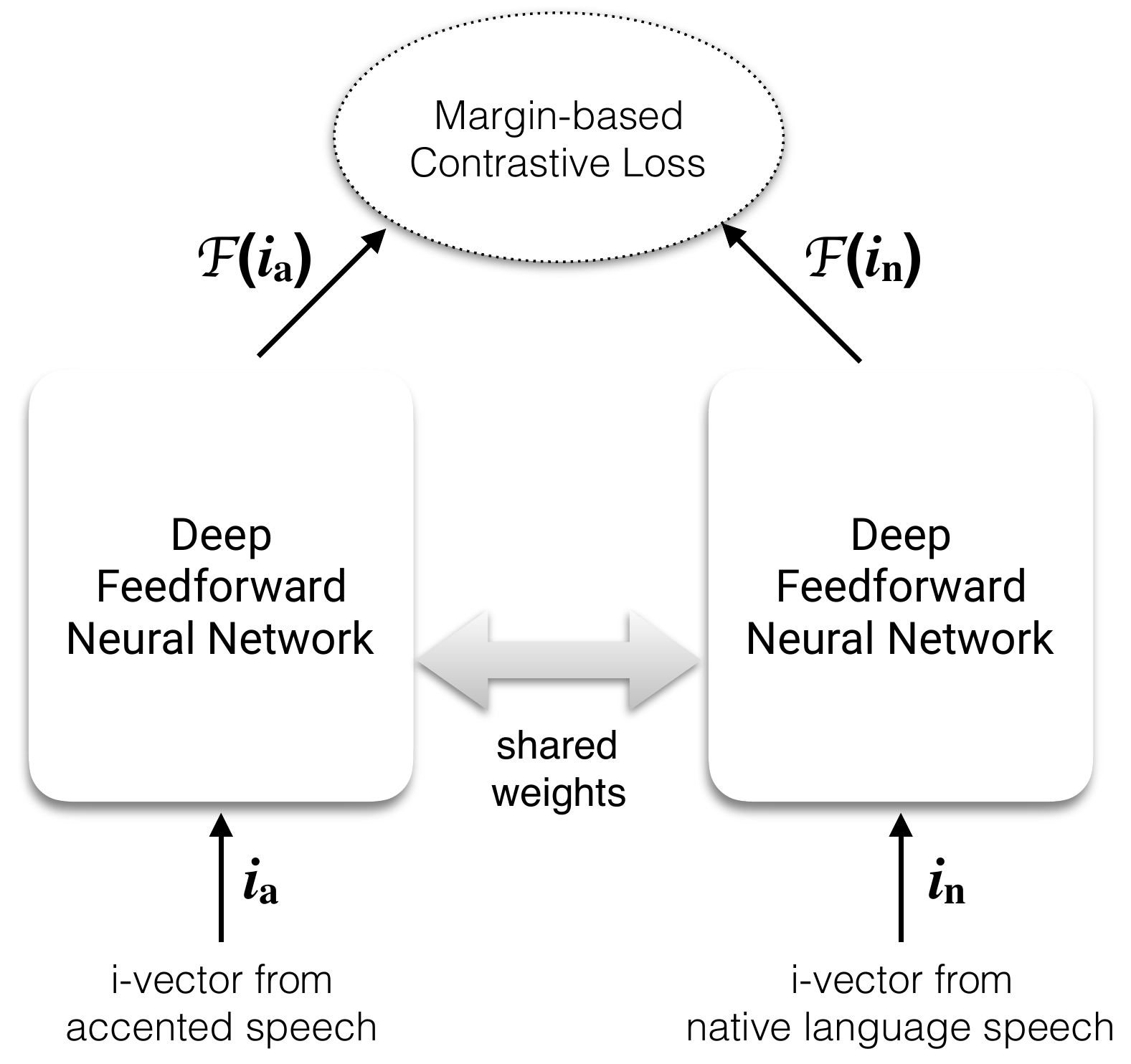}
\caption{Siamese network architecture for accent identification\label{fig:siamese}}
\end{figure}

Figure~\ref{fig:siamese} shows an illustration of the Siamese network we
used for accent identification. Each training example comprises a pair of
input i-vectors, $\{\mathbf{i}_1, \mathbf{i}_2\}$ corresponding to
an accented speech sample and a language speech sample, and a binary
label $y \in \{0,1\}$ indicating whether or not the native language
underlying the accented speech sample exactly matches that of the language
speech sample. The positive training examples correspond to accented speech 
i-vectors that are paired with native language i-vectors from the same
speaker. For the negative examples, we paired up accented speech i-vectors
with i-vectors from languages different from the one underlying the accented
speech sample. These training instances are fed to twin networks with shared
parameters which produce two outputs
$\{\mathcal{F}(\mathbf{i}_1),\mathcal{F}(\mathbf{i}_2)\}$ corresponding to
the input i-vectors $\mathbf{i}_1$ and $\mathbf{i}_2$. The whole network is
then trained to minimize the following contrastive loss function:
\begin{align} 
\mathcal{L}(\mathbf{i}_1,\mathbf{i}_2,y) &= (1-y) \cdot
\left(d(\mathcal{F}(\mathbf{i}_1),\mathcal{F}(\mathbf{i}_2))\right)^2 \nonumber \\ 
&+ y
\cdot \left(\max(0, 1-d(\mathcal{F}(\mathbf{i}_1),\mathcal{F}(\mathbf{i}_2))\right)^2
\end{align}

\noindent where $d(\cdot, \cdot)$ is a distance function between the output 
representations. 

We use a large number of positive and negative training samples to learn a distance metric between
the accented speech i-vectors and the language i-vectors. During test time,
we compare the accented speech test i-vector with a representative language
i-vector and choose the language whose i-vector representations are the nearest 
from the accented speech i-vector, according to the distance metric
learned by the Siamese network. We experiment with different strategies to
determine how the language i-vectors should be constructed during test time.
Section~\ref{sec:siameseexpts} discusses more details of these test strategies.

\section{Experimental Results}
\label{sec:expts}

\subsection{Data}
\begin{table}[t!]
\begin{tabular}{| c | c | c | c | c |}
\hline
{\textsc{Language}} & \multicolumn{3}{|c|}{Accented speech} & Native language
\\
\cline{2-4}
 & Training & Dev & Test & speech \\
 \hline
 {\textsc{BP}} & 92 & 30 & 31 & 198 \\
 {\textsc{HI}} & 66 & 22 & 22 & 206 \\
 {\textsc{FA}} & 50 & 17 & 16 & 182 \\
 {\textsc{GE}} & 55 & 18 & 18 & 161 \\
 {\textsc{HU}} & 51 & 17 & 17 & 187 \\
 {\textsc{IT}} & 34 & 11 & 12 & 168 \\
 {\textsc{MA}} & 52 & 18 & 18 & 189 \\
 {\textsc{RU}} & 44 & 14 & 14 & 172 \\
 {\textsc{SP}} & 31 & 9 & 10 & 140 \\
 {\textsc{TA}} & 37 & 13 & 12 & 128 \\
 \hline
\end{tabular}
\caption{Statistics\label{tab:data} of accented English and native language
speech data. All the displayed numbers correspond to minutes of speech.}
\end{table}
For our experiments, we used the CSLU Foreign Accented English Release 1.2
database~\cite{CSLU} that consists of telephone-quality spontaneous English
speech by native speakers of 22 different languages. We set up a 10-class
accent identification task using accented English from speakers of 10
different languages which had the most amount of data: Brazilian Portuguese (BP),Hindi
(HI), Farsi (FA), German (GE), Hungarian (HU), Italian (IT), Mandarin
Chinese (MA), Russian (RU), Spanish (SP) and Tamil (TA). For native language
speech, we used the CSLU 22 Languages Corpus~\cite{CSLU2} which contains
telephone-quality continuous speech in all the above-mentioned 10 languages.
Many of the speakers in the CSLU 22 Languages corpus also recorded speech
samples for the CSLU Foreign Accented English corpus. The samples from these speakers were used to construct positive examples for training our Siamese network. Table~\ref{tab:data} gives detailed statistics about the data used in our experiments, along with the training, development and test set splits. These splits were created using a stratified random sampler so that the proportion of different accents in each set remains almost same.

For the GMM based i-vectors, a $2048$-component UBM was trained and
$400$ dimensional i-vectors were extracted using the formulation given in Sec.~\ref{sec:factor_analysis}. These features will be referred to as \textit{GMM i-vectors} in the rest of the paper. The UBM was trained with $39$ dimensional MFCC features which were mean and variance normalized at the utterance level. The training data used in the UBM was obtained from the multilingual corpora from NIST SRE 2008 (consisting of telephone-quality speech) and the Switchboard English database~\cite{swb}. For training the DNN i-vectors, an acoustic model was developed for Switchboard using the Kaldi toolkit~\cite{Kaldi}. The DNN model generates $4677$-dimensional senone posterior features which are used in the i-vector extraction (Sec.~\ref{sec:dnn_ivec}). The i-vector training based on the DNN-UBM uses data from NIST SRE08 and Switchboard. We use $300$ dimensional i-vectors from the DNN model (henceforth referred to as \textit{DNN i-vectors}).    Both i-vectors were length normalized before the classifier training.  

\vspace{1em}
\noindent \textbf{Performance evaluation:} We used accent identification
accuracy as the primary metric to evaluate our proposed approach. This is computed
as the percentage of utterances which are correctly identified as having one
of the 10 above-mentioned accents. (A classifier based on chance would give an accuracy of $10\%$ on this task.)

\subsection{Comparing GMM i-vectors with DNN i-vectors}

\begin{table}[t!]
\centering
\begin{tabular}{| c | c | c | c | c |}
\hline
Classifier & \multicolumn{2}{|c|}{GMM i-vectors} & \multicolumn{2}{|c|}{DNN i-vectors} \\
\cline{2-5}
& Dev & Test & Dev & Test \\
\hline \hline
 LDA & 33.5 & 37.2 & 39.8 & 43.8 \\
 SVM & 35.1 & 40.2 & 39.8 & 45.2 \\
 NNET & 35.8 & 40.8 & 41.4 & 44.8 \\ 
\hline
\end{tabular}
\caption{\label{tab:gmmvsdnn} Accuracy rates (\%) from classifiers using both GMM
i-vectors and DNN i-vectors.}
\end{table}

Table~\ref{tab:gmmvsdnn} shows the performance of various baseline accent identification systems using
both GMM i-vectors and DNN i-vectors as input features. We used an LDA-based  classifier which reduces the dimensionality of the input vectors by linear projection onto a lower dimensional space that maximizes the separation between classes. We also built an SVM classifier using a radial basis function (RBF) kernel. Finally, NNET is a feed-forward neural network (with a single 128-node hidden layer) that is trained to optimize categorical cross-entropy using the Adam optimization algorithm~\cite{Kingma2014}. LDA and SVM were implemented using the scikit-learn library~\cite{scikit-learn} and NNET was implemented using the Keras toolkit~\cite{keras}. The hyper-parameters of all these models were tuned using the validation set. From Table~\ref{tab:gmmvsdnn}, we observe that the classifiers using DNN i-vectors clearly outperform the classifiers using the GMM i-vectors. This is intuitive because the DNN i-vectors carry more information about the underlying phones. Both the SVM and NNET classifiers perform comparably and outperformed the linear (LDA) classifier due to the inherent non-linearity in the data. 

In all subsequent experiments, we use the 300-dimensional DNN i-vectors (unless mentioned otherwise).

\subsection{Evaluating the Siamese network}
\label{sec:siameseexpts}
\begin{table}[t!]
\centering
\begin{tabular}{| c | c | c |}
\hline
System & Dev Accuracy & Test Accuracy \\
\hline \hline
 Siamese-1 & 42.7 & 46.4 \\ 
 Siamese-2 & 43.3 & 46.8 \\ 
 Siamese-3 & 43.3 & 47.3 \\ 
 Siamese-4 & 43.6 & 47.9 \\ 
\hline
\end{tabular}
\caption{\label{tab:siameseteststrategies} Performance (\%) of Siamese network-based classifier using different test strategies.}
\end{table}

We tried various configurations of the Siamese networks, along with varying ratios of positive and negative training examples. After tuning on the validation set, the Siamese architecture which yielded the best result consisted of 2 hidden layers with 128 nodes each and a dropout rate of 0.2 in the first hidden layer~\cite{Srivastava2014}. We used the RMSprop optimizer and the Glorot initialization for the network~\cite{Glorot2010}. The network was trained on 100,000 positive and 900,000 negative training pairs. The 10 accents were equally distributed across the positive and negative pairs.

Table~\ref{tab:siameseteststrategies} lists the accuracies of the  Siamese network-based classifiers using different strategies to choose language i-vectors during test time. We first extracted a random sample of 30 language i-vectors and computed the mean of the lowest five output scores. (Here, 30 and 5 were tuned on the validation set.) The language that was least distant from the accented test sample was chosen as the predicted output. This is the system referred to as ``Siamese-1". ``Siamese-2" refers to a system where we computed a mean language i-vector across all the i-vectors for a particular language. For ``Siamese-3", we first clustered the language i-vectors into 4 clusters using the k-means algorithm, following which we computed cluster means and chose the language whose cluster mean was minimally distant from the accented test sample. Finally, ``Siamese-4" augments ``Siamese-3" with a neural network classifier (consisting of two 8-node hidden layers with 40 input nodes and 10 output nodes). This second DNN is trained on the distance measures computed from Siamese-3 model for the 10 accent classes (4 scores for each file obtained from the four cluster mean vectors) and it predicts the target accent class. This network is trained on the validation data. In our experiments, ``Siamese-4" provided the best performance. 

Table~\ref{tab:siamese} compares the performance of our best-performing Siamese network to the two best-performing baseline systems from
Table~\ref{tab:gmmvsdnn}.  We see consistent improvements from using the
Siamese network classifier over the best baseline system on both the validation set and the test set. These improvements hold when both GMM i-vectors and DNN i-vectors are used as input features.

\begin{table}[t!]
\centering
\begin{tabular}{| c | c | c | c | c |}
\hline
Classifier & \multicolumn{2}{|c|}{GMM i-vectors} & \multicolumn{2}{|c|}{DNN i-vectors} \\
\cline{2-5}
& Dev & Test & Dev & Test \\
\hline \hline
 SVM & 35.1 & 40.2 & 39.8 & 45.2 \\
 NNET & 35.8 & 40.8 & 41.4 & 44.8 \\ 
 Siamese-4 & \textbf{37.8} & \textbf{42.3} & \textbf{43.6} & \textbf{47.9} \\ 
\hline
\end{tabular}
\caption{\label{tab:siamese} Performance (accuracy \%) of Siamese network-based classifier compared to baseline systems.}
\end{table}

\subsection{Comparison with other systems using native language i-vectors}

All the baseline systems used so far only made use of the accented English samples during training and did not make use of the native language speech samples. We compare our Siamese network-based approach with other techniques that exploit speech data from native languages during training. First, analogously to
``NNET'', we train a $2$-layer feed-forward neural network but with input
features consisting of language i-vectors concatenated with accent i-vectors. This system is referred to as ``NNET-append''. We build a second system, which we call ``NNET-nonid-twin", that is identical to the twin network Siamese architecture shown in Figure~\ref{fig:siamese}, except the weights of the twin networks are not shared. Finally, we also investigate a transfer learning based
approach, referred to as ``NNET-transfer''. For this system, we train a $2$-layer
feed-forward neural network using only language i-vectors to predict the
underlying language. Then, we use the resulting weights from the hidden
layers as an initialization for a neural network that uses accent i-vectors
as inputs to predict the underlying accent. Table~\ref{tab:nativelang}
compares these three systems with ``Siamese-4" introduced in Table~\ref{tab:siameseteststrategies}. We
observe that our proposed Siamese-network system outperforms all
the other systems which also have access to native language i-vectors.
\begin{table}[t!]
\centering
\begin{tabular}{| c | c | c |}
\hline
System & Dev Accuracy & Test Accuracy \\
\hline \hline
 NNET-append & 38.6 & 41.1 \\ 
 NNET-nonid-twin & 41.9 & 44.8 \\
 NNET-transfer & 41.7 & 45.3 \\  
 Siamese-4 & \textbf{43.6} & \textbf{47.9} \\ 
\hline
\end{tabular}
\caption{\label{tab:nativelang} Performance of various classifiers that use native language speech during training.}
\end{table}

\subsection{Accuracies on accented speech utterances with varying accent strengths}
The CSLU Foreign Accented Speech corpus contains perceptual judgments about the accents in the utterances. Each accented speech sample was independently annotated for accent strength on a scale of $1$-$4$ (where $1$ denotes a very mild accent and $4$ denotes a very strong accent) judged by three native American English speakers. Table~\ref{tab:strength} shows how the Siamese network-based classifier (``Siamese-4") performs on utterances when grouped according to accent strength. Intuitively, as the accent strength increases, our classifier accuracy increases. Despite the fact that our test set predominantly contains utterances of accent strength 3, it is interesting to see that the average accuracy on these test samples is much higher than the samples rated 1 and 2 on accent strength. 
\begin{table}[t!]
\centering
\begin{tabular}{| c | c | c | c | c |}
\hline
Accent judgment (1-4) & \% of samples & Accuracy \\
\hline \hline
1 & 10 & 34.7 \\
2 & 10 & 41.3 \\
3 & 79 & 50.4 \\
4 & 1 & 56.2 \\
\hline
\end{tabular}
\caption{\label{tab:strength}Accuracies on utterances of varying accent strengths.}
\end{table}

\subsection{System Combination}
\begin{table}[b!]
\centering
\begin{tabular}{| c | c | c | c |}
\hline
System & 1-best & 2-best & 3-best \\
\hline \hline
\multicolumn{4}{|c|}{Individual systems} \\
\hline
SVM & 45.2 & 64.9 & 73.1 \\
NNET & 44.8 & 64.1 & 72.4 \\
Siamese-4 & 47.9 & 70.2 & 80.4 \\
\hline \hline
\multicolumn{4}{|c|}{Combined systems} \\
\hline
Majority-voting & 48.6 & 73.3 & 85.1 \\
Weighted-voting & \textbf{48.8} & \textbf{75.1} & \textbf{87.4} \\
\hline
\end{tabular}
\caption{\label{tab:fusion} N-best accuracies (\%) on the test set from both individual systems and combined systems.}
\end{table}
Table~\ref{tab:fusion} shows the accuracies from our two top baseline systems and our best Siamese network based system, Siamese-4, when the correct accent is among the top 2 and top 3 accent predictions. We observe that the accuracies dramatically improve across all three individual systems when moving from 1-best to 2-best and 3-best accuracies. This is indicates that a significant part of the confusions seen across the classes are confined to the top 3 predictions. 

We also combine the outputs from the three individual systems. We adopt a simple majority voting strategy: Choose the prediction that 2 or more systems agree upon. If all three systems predict different accent classes, then choose the accent predicted by Siamese-4. We also use a weighted voting strategy. The outputs from Siamese-4 were converted into posterior probabilities using a softmax layer. A weighted combination of the probabilities from SVM, NNET and Siamese-4 was used to determine the accent with the highest probability. (Weights of 0.3, 0.3 and 0.4 were used for SVM, NNET and Siamese-4, respectively.) The N-best accuracies for both these combined systems are shown in Table~\ref{tab:fusion}. As expected, we observe performance gains from combining the outputs of all three individual systems; the gains are larger in the 2-best and 3-best cases.

\section{Discussion}
\label{sec:discussion}

\begin{figure}[t!]
\centering
\includegraphics[width=0.9\linewidth]{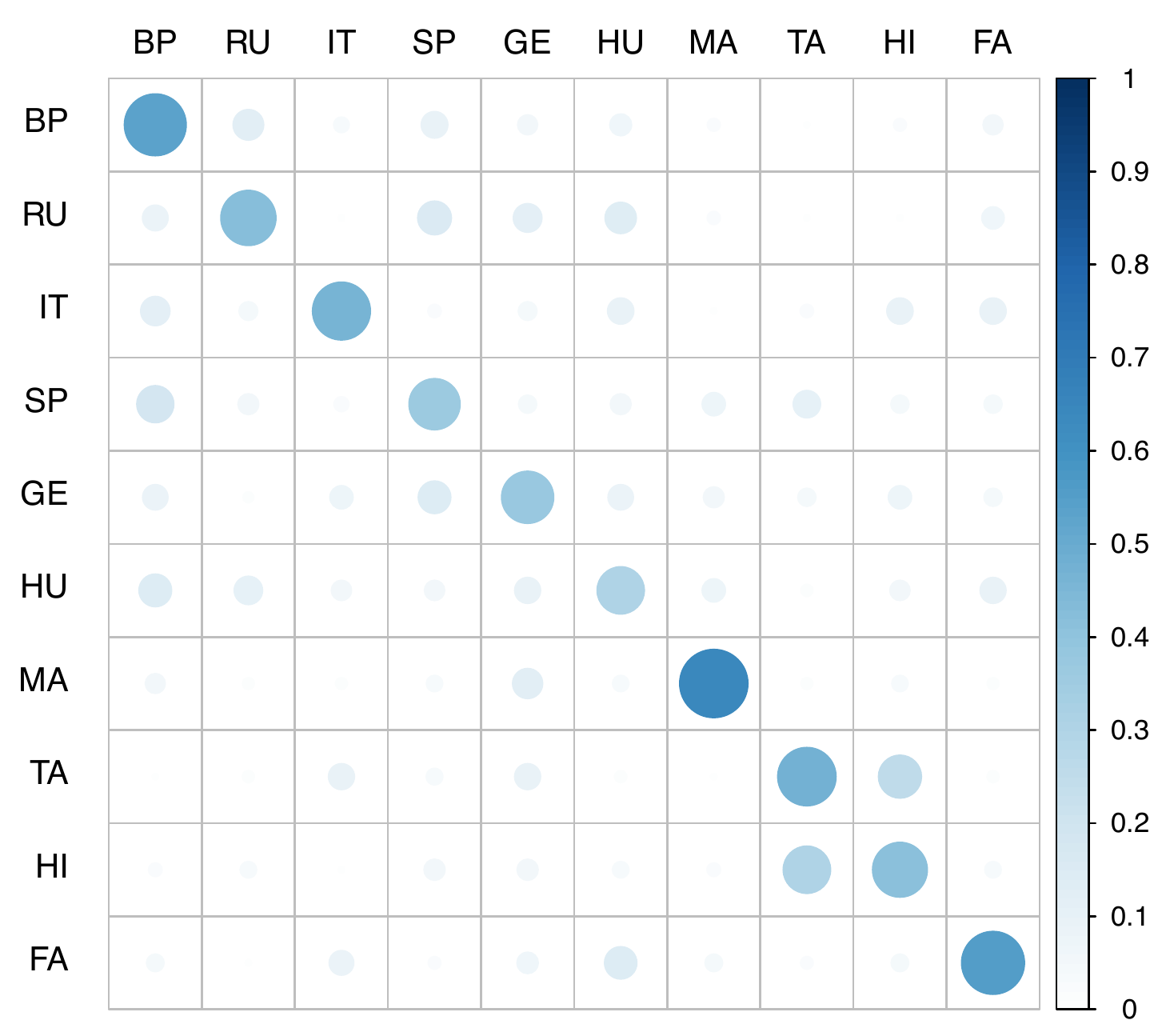}
\caption{\label{fig:bubble1} Bubble plot visualizing the confusion matrix of test set predictions from Siamese-4.}
\end{figure}

\begin{figure}[h!]
\centering
\includegraphics[width=0.9\linewidth]{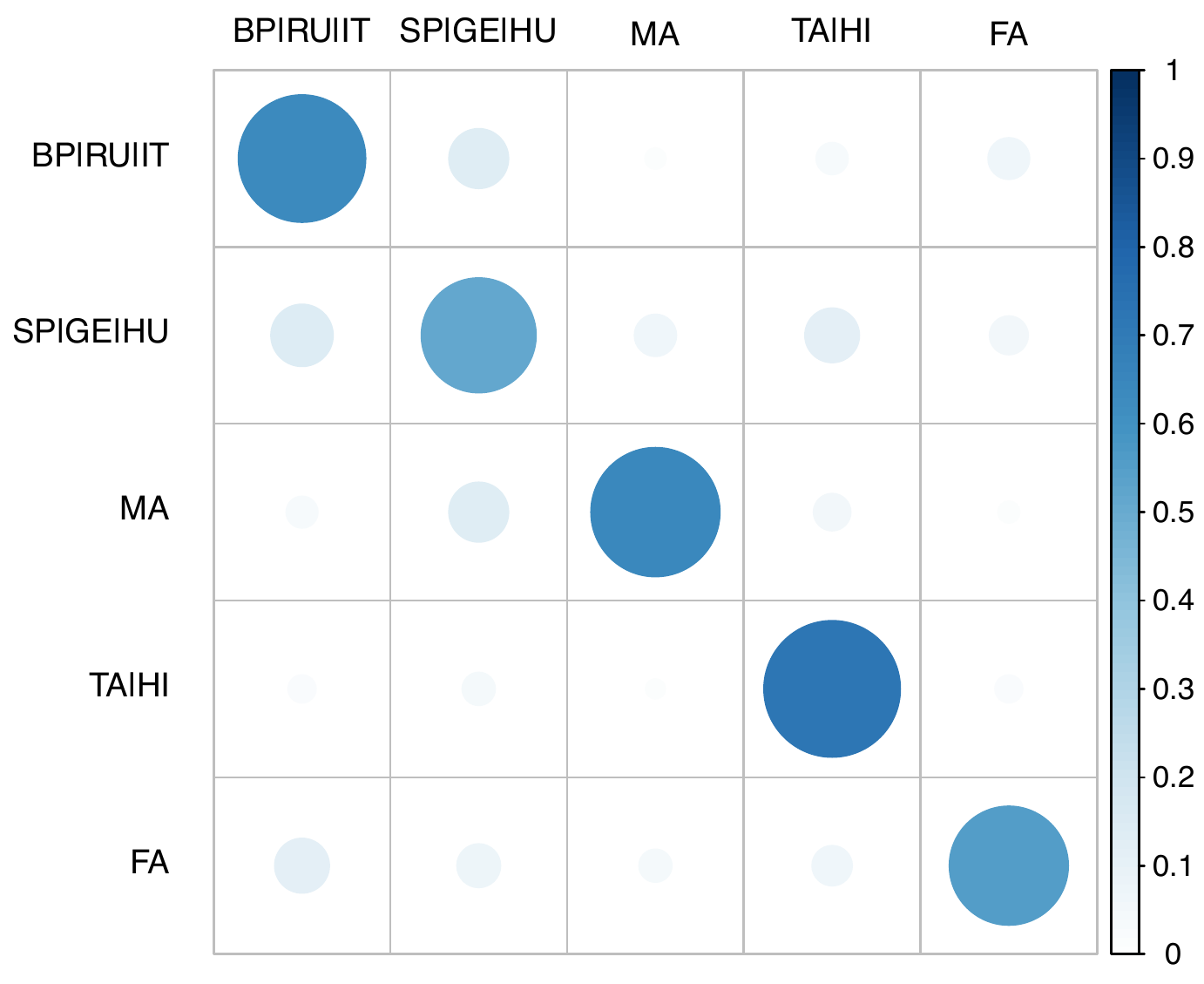}
\caption{\label{fig:bubble2} Bubble plot visualizing the confusion matrix of test set predictions from Siamese-4, after grouping related languages.}
\end{figure}

It is illustrative to visualize the accent predictions made by our proposed Siamese network-based classifier in order to learn more about the main confusions that are prevalent in the system. We visualize the confusion matrix of the Siamese-4 classifier on the test samples using a heat map as shown in Figure~\ref{fig:bubble1}. Darker/bigger bubbles against a column for a given row indicate a larger number of samples were predicted as being of the accent labeled against the column. For each accent class, Figure~\ref{fig:bubble1} shows there are a sizable number of test examples that are correctly predicted as evidenced by the dark bubbles along the diagonal. We also see more than one darkly-colored bubble along the non-diagonal indicating strong evidence for confusion. For example, Hindi-accented English samples are often confused as being Tamil-accented and conversely, Tamil-accents in some test samples are mistaken for Hindi-accents. This is very intuitive given that both these accents are very closely related and correspond to the same geographical region. Indeed, if we group the languages according to the language families that they belong to, i.e. \{BP, RU and IT\}, \{SP, GE and HU\}, \{MA\}, \{HI and TA\} and \{FA\}, the corresponding confusion matrix exhibits far less confusion as shown in Figure~\ref{fig:bubble2}.


\section{Conclusions}
In this paper, we explore the problem of accent identification from non-native English speech. We propose a novel approach based on deep Siamese neural networks that uses i-vectors extracted from both accented speech and native language speech samples and learns a semantic distance between these feature representations. On a 10-class accent identification task, our proposed approach outperforms a neural network-based classifier using both GMM-based i-vectors and DNN-based i-vectors with relative accuracy improvements of 15.4\% and 7.0\%, respectively.

In this work, we focused on the acoustic characteristics of an accent induced by a speaker's native language. Accents are also correlated with specific lexical realizations of words in terms of pronunciations and variations in word usage and grammar. As future work, we plan to explore how to incorporate the pronunciation model and language model based features to automatic identification of speech accents.

\section{Acknowledgements}
The authors thank the organizers of the MSR-ASI workshop, Monojit Choudhury, Kalika Bali and Sunayana Sitaram for  the opportunity and all their help, as well as the other project team members Abhinav Jain, Gayatri Bhat and Sakshi Kalra for their support. We also gratefully acknowledge the logistical support from Microsoft Research India (MSRI) for this project as well as access to Microsoft Azure cloud computing resources. 

\newpage
\bibliographystyle{IEEEbib}
\bibliography{refs}

\end{document}